\documentclass{article}
\usepackage[numbers,sort]{natbib}
\usepackage{spconf,amsmath,graphicx,setspace}
\usepackage{multirow}
\usepackage{bbm} 				
\usepackage{multirow} 			
\usepackage{booktabs} 			
\usepackage{color}
\usepackage{amssymb}			
\usepackage{enumitem}			
\usepackage{bm}
\usepackage{tabularx}

\title{FEATURE ADVERSARIAL DISTILLATION FOR POINT CLOUD CLASSIFICATION}
\name{YuXing Lee, Wei Wu}
\address{Department of Computer Science, Inner Mongolia University, China}

\begin{document}
%
\maketitle
\begin{abstract}
Due to the point cloud's irregular and unordered geometry structure, conventional knowledge distillation technology lost a lot of information when directly used on point cloud tasks.
In this paper, we propose Feature Adversarial Distillation (FAD) method, a generic adversarial loss function in point cloud distillation, to reduce loss during knowledge transfer.
In the feature extraction stage, the features extracted by the teacher are used as the discriminator, and the students continuously generate new features in the training stage. 
The feature of the student is obtained by attacking the feedback from the teacher and getting a score to judge whether the student has learned the knowledge well or not.
In experiments on standard point cloud classification on ModelNet40 and ScanObjectNN datasets, our method reduced the information loss of knowledge transfer in distillation in 40x model compression while maintaining competitive performance. 
\end{abstract}

\begin{spacing}{1} 
\section{Introduction}
\label{sec:intro}
With the development of deep neural networks (DNNs), deep learning has made rapid progress in speech, computer vision, and natural language processing.
In recent years, the technology of deep learning has been extended to the unordered and irregular data of point clouds. 
The pioneering work is PointNet \cite{PointNet}, a full MLP-based method, which completed various downstream tasks such as point cloud classification, semantic segmentation, and part segmentation, and obtained a promising result.
However, it only pays attention to the global features of the point cloud and cannot capture the local features of the point cloud. 
Thus, PointNet++ \cite{pointnet++} is proposed to sample the point cloud by grouping and then use PointNet on each group of point clouds to capture the local features. 
Afterward, the graph-based method \cite{DGCNN}\cite{DeepGCN}\cite{inner_group} has been put forward, such as DGCNN \cite{DGCNN} and DeepGCN \cite{DeepGCN} series network. 
DGCNN first proposed EdgeConv, which uses the relationship between the vertices and edges in point cloud data to improve performance. 
DeepGCN introduces a series of methods from conventional images to the task of point clouds like ResGCN, DenseGCN, and Dilated Convolution in GCN corresponding to the classical network architecture ResNet \cite{ResNet}, DenseNet \cite{DenseNet} and Dilated Convolution \cite{DilatedConvolution}.
This contribution makes the network structure of the point cloud as flexible as the image. 

Recently, Transformer \cite{Transformer} has been widely used in image tasks such as image classification \cite{VisioTransformer}, object detection, and semantic segmentation \cite{Segmenter}. 
Point clouds are unordered and irregular data, and the transformer method in the image can not be directly applied to the point cloud tasks. 
Guo proposed the point cloud transformer (PCT) \cite{Pct} and Zhao proposed the point transformer (PT) \cite{PointTransformer} for the point cloud task with transformer. 
PT introduced vector attention \cite{VectorAttention} into the transformer for the first time and achieved remarkable results in various downstream tasks. 
PCT proposed Neighbor Embedding and Offset Attention and applied them to the Encoder of the transformer. 
Due to the complexity of the above model, Ma rethought the point cloud network in MLP and proposed PointMLP \cite{PointMLP}. 
Its most important contribution is that the lightweight geometric affine and ResP model \cite{PointMLP} are introduced, which makes the PointMLP network more flexible. 
However, the parameters of these models are very large, which is difficult to deploy and maintain efficiently in resource-limited scenarios and devices.

\begin{figure*}[!t]
	\centering
	\includegraphics[width=6in]{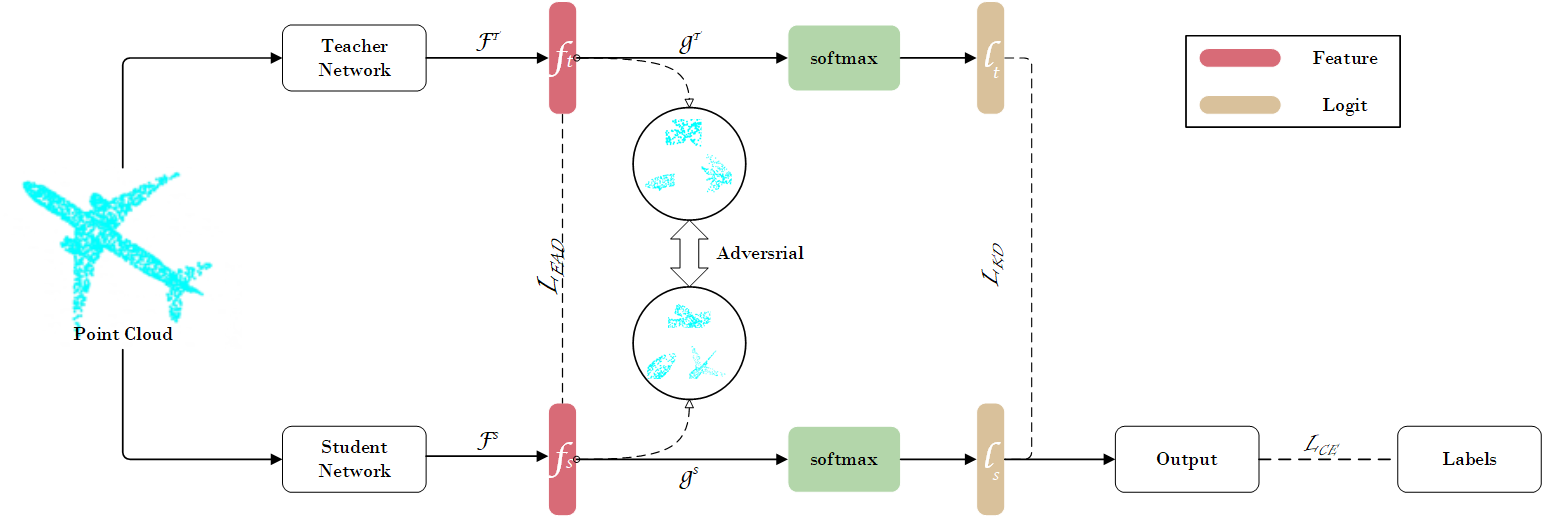}
	\caption{Overview of the proposed method.}
	\label{fig1}
\end{figure*}

Knowledge distillation is a technology means to effectively solve point cloud tasks in resource-limited scenarios.
Hinton first proposed the technology of knowledge distillation \cite{HintonKD}, 
which introduces the teacher network, a pre-trained model to promote the student network by the soft probability distribution.
After that, researchers improved distillation technology in various aspects, such as soft predictions \cite{fitnets}, intermediate representations \cite{relational_KD}, attention maps \cite{AttentionDistillation}, etc.
But these methods are not effective when used directly on the point cloud, due to the point cloud structure is irregular and unordered.

Recently, some scholars have begun to study the compression technology of point clouds or point cloud LiDAR data to explore lightweight solutions \cite{SA-LPCC} \cite{HEVC}.  
The earlier work was to improve the performance of point cloud distribution by introducing the loss function of open set recognition \cite{bhardwaj2021empowering}.
But the work has a shortcoming in that the distillation model is PointNet, which lacks flexibility and makes it difficult to do large-scale model compression.
PointDistiller \cite{pointdistiller} employs local distillation, which encodes the local geometric structure of point clouds with dynamic graph convolution.
The Point-to-voxel (PVD) \cite{point_to_voxel} is used to compress and distill the model by converting point cloud mapping to voxels.
However, these methods are complex and lack universality.

We are the first to study point cloud data distillation that is generalizable and can be applied to various point cloud distillation tasks. 
In this work, we propose Feature Adversarial Distillation (FAD), a simple and effective method for point cloud distillation, which does not require voxelization or dynamic graph convolution to distill the point cloud network and has a higher generalization ability in point cloud tasks. 
In the feature extraction stage, the features extracted by the teacher are used as the discriminator, and the students continuously generate new features in the training stage. 
The feature of the students is obtained by attacking the feedback from the teacher and getting a score to judge whether the students have learned the knowledge well or not. 
By distilling models with different architectures, we conducted experiments on standard point cloud classification datasets ModelNet40 and ScanObjectNN. 
In the MLP-based method, we compress the model by 40 times. 
Our method improves the accuracy of the compressed model to $91.65\%$. 
In the graph-based method, we compress the model by 4 times. 
On the conventional KD loss function, the experimental accuracy is $88.17\%$, and after using our proposed FAD loss function, the accuracy is $90.51\%$. 
Experimenting on the real-world dataset ScanObjectNN also confirms the effectiveness of our model. 
By compressing PointMLP 40 times, we still improve the accuracy from $75.60\%$ to $76.27\%$ compared with the conventional KD loss function. 
In addition, we also study the performance on heterogeneous networks, in which the teacher model and student model use different methods.






\section{METHODOLOGY}
\label{sec:pagestyle}
In this section, we will describe our method in detail.
The overview of the proposed method is shown in \textbf{Fig. 1}.
The point cloud input into the teacher and student network respectively.
The features of teachers and students are obtained by extracting features.
The knowledge gained by students can compete with teachers by attacking and the teacher feedback to the student a score, telling students that the knowledge learned is well or not.
At this time, students need to learn from teachers to reduce the differences in knowledge between students and teachers.
The KD loss function is calculated by the soft labels of softmax and using the hard label to calculate the cross-entropy loss function.
Then we combine the FAD, KD, and CE to establish a joint loss function for training.

{
\setlength{\parindent}{0cm}
\textbf{Distillation detail:}
Let $F^{S}$ and $F^{T}$ be the functions denoting the student's feature extraction and the teacher's feature extraction.
The output is the student and teacher features distribution, noted $f_{S}$ and $f_{T}$.
Following we have the production layer of teacher and student, which is the fully connected layer actually, to compute its soft labels with defined $l_{S}$ and $l_{T} $, 
whose functions are denoted by $g^{S}$ and $g^{T}$. It can be written as:
}

\begin{equation}
  f_{S} = F^{S}(x)
\end{equation}
\begin{equation}
  f_{T} = F^{T}(x)
\end{equation}
\begin{equation}
  l_{S} = g^{S}(f_{S})
\end{equation}
\begin{equation}
  l_{T} = g^{T}(f_{T})
\end{equation}

In final fully connected layer, we need to calculate cross-entropy loss, noted $L_{CE}$ which is shown in Eq.5 and 6.

\begin{equation}
  L_{CE} = \left(\sigma\left(l_{S}\right);y\right) 
\end{equation}
where $\sigma$ is softmax activation function and y is the true label in dataset.
\begin{equation} 
  \sigma\left(l_{S}\right)=\frac{\exp l_{i}}{\sum_{j=1} \exp l_{j}} 
\end{equation}
where is the summary of all the classes.

In logits layer, we use KD technology to minimize the following loss function.
As shown in Eq.7.

\begin{equation} 
  L_{KD} = L_{CE}\left(\sigma\left(l_{S};\tau_{K}\right);\sigma\left(l_{T};\tau_{K}\right)\right) 
\end{equation}
where $\tau$ is the temperature hyperparameter used to modified the softmax function as shown in Eq.8, 
it was first proposed by \cite{HintonKD}.
\begin{equation}
  \sigma\left(l_{S};\tau_{K}\right)=\frac{\exp\left(l_{i}/\tau_{K}\right)}{\sum_{j=1}\exp\left(l_{j}/\tau_{K}\right)}
\end{equation}
where the summation of all the classes by temperature parameter.

In feature layer, we calculate the feature adversarial loss by FAD.
We propose an abstract representation method in feature adversarial, which can be written as three formulas [9-11]: 


\begin{equation} 
  L_{FAD} = \sum_{i=1}^{N} \lvert \max\left(f_{Tij}\right) - \max \left(f_{Sij}\right) \rvert \left(j \in 1 \thicksim D \right)
\end{equation}
where N is the number of points with D dimensions in the input point cloud.
In Eq.10, it means finding the max feature at each point in dimensions and calculating the subtraction for both the teacher and the student.
Then take the absolute value and summary them.

\begin{equation} 
  L_{FAD} = \sum_{i=1}^{N} \lvert \min \left(f_{Tij}\right) - \min \left( f_{Sij} \right) \rvert \left(j \in 1 \thicksim D \right)
\end{equation}
which means finding the smallest feature for each point in the dimension and subtracting, where N is the number of points with D dimensions in the input point cloud.
Then take the absolute value and summary them.

\begin{equation}
  L_{FAD} = \frac{1}{ND}\sum_{i = 1}^{N}\sum_{j = 1}^{D} \left\lvert f_{Tij}- f_{Sij}\right\rvert 
\end{equation}
where N is the number of points with D dimensions in the input point cloud.
Calculation the difference between each dimension of each point by subtraction. 
Then sum and average, where the average is $\frac{1}{ND} $, them to represent the feature described in the point cloud.
Finally, the student's and teacher's features are subtracted and taken as absolute values.

{
\setlength{\parindent}{0cm}
In summary our loss function is:
}

\begin{equation}
  L = \alpha L_{FAD}+ \beta L_{KD}+ \gamma L_{CE}
\end{equation}
where $\alpha$, $\beta$, $\gamma$ is hyperparameter of the weight of each part loss and $\alpha + \beta + \gamma = 1$

\begin{table}[htb]
    \vspace{-0.5cm}
    \centering 
    \caption{Parameters and FLOPs of the teacher and the student network}
    \setlength{\tabcolsep}{4mm}{
    \begin{tabular}{c | c | c | c}
    \hline
      Model   &      Type      &  Param    &  FLOP\\
    \hline
    \multirow{2}{*}{PointMLP} & \multirow{1}{*}{teacher} & \multirow{1}{*}{13.2M} & \multirow{1}{*}{503.49G} \\ \cline{2-4}
                              & \multirow{1}{*}{student} & \multirow{1}{*}{0.3M} & \multirow{1}{*}{20.51G} \\ \cline{2-4}
    \hline 
    \multirow{2}{*}{ResGCN} & \multirow{1}{*}{teacher} & \multirow{1}{*}{5.3M} & \multirow{1}{*}{127.76G} \\ \cline{2-4}
                            & \multirow{1}{*}{student} & \multirow{1}{*}{1.7M} & \multirow{1}{*}{30.34G} \\ \cline{2-4}
    \hline
    \end{tabular}}
\end{table}

\section{EXPERIMENTS}
\label{sec:typestyle}

{
\setlength{\parindent}{0cm}
\vspace{-0.1cm}
\textbf{Dataset:}
We conduct experiments on two standard point cloud classification datasets, \emph{i.e.}, ModelNet40 \cite{ModelNet40} and ScanObjectNN \cite{ScanObjNN}.
ModelNet40, a CAD Model dataset, has 9843 training samples and 2468 test samples including 40 object categories, which collects 1024 3-dimensional points utilizing random translation for each sample.
ScanObjectNN is a real-world object dataset containing 15K objects in 15 categories, which collects 1024 3-dimensional points utilizing random translation for each sample.
}
~\\
{
\setlength{\parindent}{0cm}
\textbf{Experimental scheme:}
The parameters and FLOPs compared with the teacher model and student model are shown in \textbf{Table 1}. 
The parameters of the student model are 4x compressed by the teacher model from the graph-based method. 
For the MLP-based method, the student model is 40x compressed by the teacher model parameters. 
To prove our method's efficiency, we try the following schemes on the ModelNet40 dataset: homogeneous networks and heterogeneous networks. 
Homogeneous networks mean the teacher model is the MLP-based method and the student model is the MLP-based method, or the teacher model is the graph-based method and the student model is the graph-based method. 
Heterogeneous networks mean that the teacher model is the MLP-based method and the student model is the graph-based method, or the teacher model is the graph-based method and the student model is the MLP-based method. 
We are also experimenting with a more difficult real-world dataset, ScanObjectNN. 
Finally, we conduct ablation studies for the proposed three different methods to represent FAD in the real-world dataset ScanObjectNN. 
The architecture of our experiments is shown in \textbf{Fig. 2}.
}

\begin{figure}[htb]
  \vspace{-0.3cm}
  \begin{minipage}[b]{1\linewidth}
    \centering
    \centerline{\includegraphics[width=8.5cm]{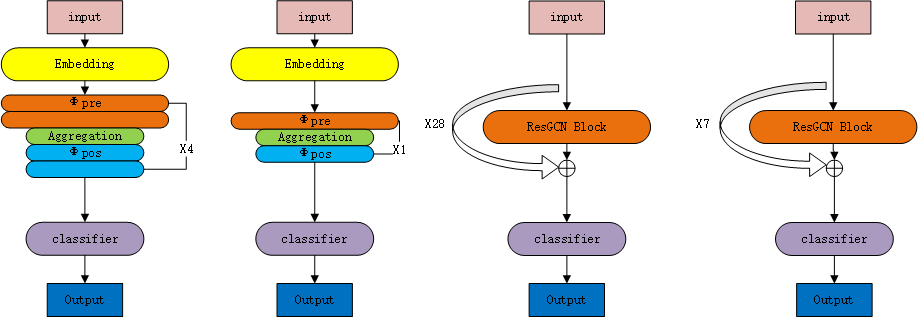}}
  \end{minipage}
  \caption{Distillation model architecture. Left is PointMLP teacher model and student model architecture. Right is ResGCN teacher model and student model architecture.}
  \label{fig:res}
\end{figure}

{
\setlength{\parindent}{0cm}
\vspace{-0.2cm}
\textbf{Implementation details:}
To validate our method's efficiency, we use the same parameters in each experiment.
We trained on a single GPU for 200 epochs, using PyTorch as the framework and SGD as the optimizer, with 0.9 of momentum and 0.0002 of weight decay.
The initial learning rate is 0.01, and the batch size is 32.
The knowledge distillation temperature is set to 4.
}

\begin{table}[htb]
  \centering 
  \caption{Distillation for ModelNet40}
  \setlength{\tabcolsep}{3mm}{
  \begin{tabular}{c | c | c | c}
  \hline
    Teacher:baseline&           Student &      Norm KD      &  FAD    \\
  \hline
  \multirow{2}{*}{PointMLP:93.73} & \multirow{1}{*}{PointMLP} & \multirow{1}{*}{89.27} & \multirow{1}{*}{\textbf{91.65}} \\  \cline{2-4}
                    & \multirow{1}{*}{ResGCN-7} & \multirow{1}{*}{90.11} & \multirow{1}{*}{\textbf{91.93}} \\  \cline{2-4}
  \hline 
  \multirow{2}{*}{ResGCN:92.38} & \multirow{1}{*}{PointMLP} & \multirow{1}{*}{89.42} & \multirow{1}{*}{\textbf{91.61}} \\ \cline{2-4}
                    & \multirow{1}{*}{ResGCN-7} & \multirow{1}{*}{88.17} & \multirow{1}{*}{\textbf{90.51}} \\ \cline{2-4}
  \hline
  \end{tabular}}
\end{table}

\begin{table}[htb]
    \vspace{-0.5cm}
    \centering 
    \caption{Compare with other typical Knowledge Distillation method in ModelNet40 dataset}
    \setlength{\tabcolsep}{2mm}{
    \begin{tabular}{c | c | c | c}
    \hline
      Model   & Type  & Method &  Overall Accuracy \\
    \hline
    \multirow{1}{*}{PointMLP} & \multirow{1}{*}{teacher} & \multirow{1}{*}{-} & \multirow{1}{*}{93.64$\%$} \\ \cline{2-4}
                              
    \hline 
    \multirow{5}{*}{PointMLP} & \multirow{4}{*}{student} & \multirow{1}{*}{KD} & \multirow{1}{*}{89.27$\%$} \\ \cline{3-4}
                            &  & \multirow{1}{*}{RKD \cite{relational_KD}} & \multirow{1}{*}{89.62$\%$} \\ \cline{3-4}
                            &  & \multirow{1}{*}{FitNet \cite{fitnets}} & \multirow{1}{*}{90.14$\%$} \\ \cline{3-4}
                            &  & \multirow{1}{*}{FitNet+KD} & \multirow{1}{*}{90.72$\%$} \\ \cline{3-4}
                            &  & \multirow{1}{*}{FAD} & \multirow{1}{*}{\textbf{91.65$\%$}} \\ \cline{3-4}
    \hline
    \end{tabular}}
  \end{table}
\vspace{-0.2cm}
\subsection{ModelNet Classification}
\label{ssec:subhead}
We compare the differences between homogeneous networks (\emph{i.e.}, the teacher network is an MLP-based method and the student network is an MLP-based method) and heterogeneous networks (\emph{i.e.}, the teacher network is an MLP-based method and the student network is a graph-based method). 
The overview of our experiment is shown in \textbf{Table 2}.
In the homogeneous network, we can see in the MLP-based method that the norm KD accuracy is \textbf{89.27$\%$} and
FAD accuracy is \textbf{91.65$\%$} compared with the teacher's accuracy, which reduces loss in knowledge transfer from \textbf{4.46$\%$} to \textbf{2.08$\%$}.
In the graph-based method, FAD and norm KD compared with the teacher's accuracy reduce loss in knowledge transfer from \textbf{4.21$\%$} to \textbf{1.87$\%$}.
In the heterogeneous network, when the teacher is the MLP-based method and the student is the graph-based method, FAD loss falls \textbf{3.62$\%$} to \textbf{1.8$\%$}.

We compared other typical knowledge distillation technologies as shown in \textbf{Table 3}, 
where RKD is a typical representative of relational distillation and FitNet is a representative of characteristic distillation. 
Finally, we also compared the combination of characteristic distillation and logit distillation.
Our FAD achieves the state-of-the-art.

\textbf{Fig. 3} displays the valuation compared with FAD and KD.
The left figure show the accuracy in training and testing. 
FAD is more stable than conventional KD in testing.
The right figure show the loss in training and testing. 
The loss of FAD is less than that of KD.
\begin{table}
\vspace{-0.5cm}
  \centering 
  \caption{Distillation for ScanObjectNN}
  \setlength{\tabcolsep}{6.5mm}{
  \begin{tabular}{c | c | c }
  \hline
    Model   &      Loss Function      &  Accuracy    \\
  \hline
  \multirow{1}{*}{Teacher} & \multirow{1}{*}{CE}      & \multirow{1}{*}{85.88} \\ \cline{2-3}
  \hline 
  \multirow{2}{*}{Student} & \multirow{1}{*}{Norm KD} & \multirow{1}{*}{75.60} \\ \cline{2-3}

                           & \multirow{1}{*}{FAD}     & \multirow{1}{*}{\textbf{76.27}} \\ \cline{2-3}

  \hline
  \end{tabular}}
\end{table}

\begin{figure}[htb]
    \vspace{-0.4cm}
    \begin{minipage}[b]{.48\linewidth}
      \centering
      \centerline{\includegraphics[width=4.0cm]{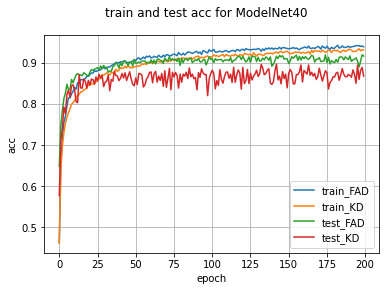}}
    \end{minipage}
    \hfill
    \begin{minipage}[b]{0.48\linewidth}
      \centering
      \centerline{\includegraphics[width=4.0cm]{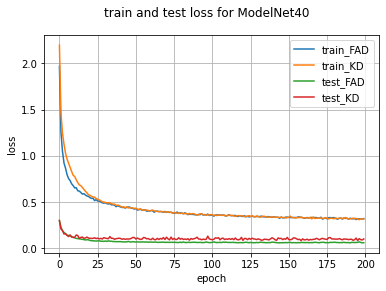}}
    \end{minipage}
    \caption{The detail for training and testing in accuracy and loss.}
    \label{fig:res}
\end{figure}
\vspace{-0.2cm}
\subsection{ScanObjectNN Classification}
In practical applications, the majority of data is real-world objects.
Thus, we test the ScanOjbectNN , a real-world object dataset.
The result is shown in \textbf{Table 4}. 
We can see that it is more difficult to transfer knowledge through knowledge distillation when using real-world datasets.
Conventional KD loss will lose a large amount of feature information, which reduces 10.28$\%$ in accuracy, 
but our FAD loss reduces it from \textbf{10.28$\%$} to \textbf{9.61$\%$}.
\vspace{-0.2cm}
\subsection{Ablation Studies}
In section $2$, we talk about different methods to describe FAD.
In this section, we compared these methods to the results in the ModelNet40 dataset, as shown in \textbf{Table 5}. 
We can see that the accuracy of Norm KD is 89.27$\%$, with MAX (Eq.10) its accuracy is 91.29$\%$, with MIN (Eq.11) its accuracy is 91.23$\%$, and with MEAN (Eq.12) the accuracy rate is 91.65$\%$.
We can see improvement in accuracy regardless of whether we use MIN, MAX, or MEAN to describe FAD.
When using \textbf{the MEAN} to describe the FAD, it achieves the best performance of \textbf{91.65$\%$}.

\begin{table}
    \vspace{-0.5cm}
  \setlength\parindent{0em}
  \centering 
  \caption{Compare with different methods to describe FAD}
  \setlength{\tabcolsep}{6.5mm}{
  \begin{tabular}{c | c | c}
  \hline
    Model   &      Loss Function     &  Accuracy    \\
  \hline
  \multirow{1}{*}{Teacher} & \multirow{1}{*}{CE} & \multirow{1}{*}{93.73} \\ \cline{2-3}
  \hline 
  \multirow{4}{*}{Student} & \multirow{1}{*}{Norm KD} & \multirow{1}{*}{89.27} \\ \cline{2-3}

                           & \multirow{1}{*}{MIN} & \multirow{1}{*}{91.23} \\ \cline{2-3}
                           & \multirow{1}{*}{MAX} & \multirow{1}{*}{91.29} \\ \cline{2-3}
                           
                           & \multirow{1}{*}{MEAN} & \multirow{1}{*}{\textbf{91.65}} \\ \cline{2-3}
  \hline
  \end{tabular}}
\end{table}
\end{spacing}

\vspace{-0.4cm}
\section{CONCLUSION}
\vspace{-0.2cm}
\label{sec:majhead}
In this paper, a sample and effective knowledge distillation method for point cloud is introduced, Feature Adversarial Distillation (FAD), which can reduce feature loss in point cloud knowledge distillation. 
We adopt the average to express point cloud feature and build a new loss function based on adversarial. 
Establishing a joint loss function by $L_{FAD}$, $L_{KD}$, and $L_{CE}$, the teacher can use their feature knowledge to supervise student learning, which can provide a score, and the students update their knowledge and adversarial with the teacher continuously. 
Our study has shown that in CAD models like the ModelNet40 dataset, FAD can well transfer knowledge and have high performance. 
Especially in real-world data, FAD can reduce the loss in knowledge transfer efficiently.

\vspace{-0.5cm}
\section{Acknowledge}
This work is supported by the Inner Mongolia Science and Technology Project No.2021GG0166 and National Natural Science Foundation of China project No.61763035

\vfill\pagebreak

{
  \bibliographystyle{IEEEbib}
  \begin{spacing}{0} 
    \bibliography{ref}
  \end{spacing} 
}

\end{document}